\documentclass[11pt]{article}

\usepackage[preprint]{acl}

\usepackage{times}
\usepackage{latexsym}

\usepackage[T3, T1]{fontenc}

\usepackage[utf8]{inputenc}

\usepackage{tipa}
\usepackage{tabularx}

\usepackage{microtype}

\usepackage{inconsolata}

\usepackage{graphicx}
\usepackage{booktabs}
\usepackage{amsmath}
\usepackage{multirow}
\usepackage{tipa}

\usepackage{graphicx}
\usepackage{amsmath}
\usepackage{textgreek}
\usepackage{tikz}

\usetikzlibrary{arrows.meta,shapes.geometric,shapes.symbols,shadows}

\usetikzlibrary{shapes, arrows, positioning, calc}
\usepackage{pgfplots}
\pgfplotsset{compat=1.17}
\usepackage{subcaption} 
\usepackage{microtype}
\usepackage[utf8]{inputenc}
\usepackage{newunicodechar}
\newunicodechar{̂}{\^}

\title{Kunnafonidilaw ka Cadeau: an ASR dataset of present-day Bambara}

\author{
 \textbf{Yacouba Diarra},
 \textbf{Panga Azazia Kamaté},\\
 \textbf{Nouhoum Souleymane Coulibaly},
  \textbf{Michael Leventhal}\\
 RobotsMali AI4D Lab, Bamako, Mali;
  \small{
    \textbf{Correspondence:} \href{mailto:research@robotsmali.org}{research@robotsmali.org}
  }
}
\begin{document}
\maketitle
\begin{abstract}
We present Kunkado, a 160-hour Bambara ASR dataset compiled from Malian radio archives to capture present-day spontaneous speech across a wide range of topics. It includes code-switching, disfluencies, background noise, and overlapping speakers that practical ASR systems encounter in real-world use. We finetuned Parakeet-based models on a 33.47-hour human-reviewed subset and apply pragmatic transcript normalization to reduce variability in number formatting, tags, and code-switching annotations. Evaluated on two real-world test sets, finetuning with Kunkado reduces WER from 44.47\% to 37.12\% on one and from 36.07\% to 32.33\% on the other. In human evaluation, the resulting model also outperforms a comparable system with the same architecture trained on 98 hours of cleaner, less realistic speech. We release the data and models to support robust ASR for predominantly oral languages.
\end{abstract}

\section{Introduction}
\label{sec:intro}
Data availability for automatic speech recognition (ASR) for the Bambara language has increased significantly this year. For about three years, Jeli-ASR \cite{Diarra2022Griots}, a corpus of 30 hours of transcribed griot narrations, had been the only open ASR dataset for Bambara, but in late 2025, a team at RobotsMali AI4D Lab released a 612 hour dataset as part of the African Next Voices (ANV) project \cite{diarra2025dealinghardfactslowresource}, scaling the amount of open data available by a factor of $20$x. In both cases, the audio was recorded during the project in a relatively controlled environment with consistent quality control prior to transcription.

Cost and the challenges of field collection have led to many initiatives aiming to increase speech data for low resource languages (LRLs) to either align data that has already been recorded such as Bible readings (\citealp{cmuwilderness_black}; \citealp{pratap2023scalingspeechtechnology1000}) or employ synthetic generation techniques \cite{derenzi2025syntheticvoicedataautomatic}. 
The former requires a pretrained acoustic model to produce alignments between acoustic features and the corresponding phonemes in a long transcript \cite{asr-pipeline-for-lrl-pomak} or an existing aligned set to train a speech synthesizer to generate phones from the transcript and find those phones in the audio \cite{cmuwilderness_black}.
Synthetic speech data can be useful for robustness training but its value is significantly lower than that of real speech. It also increases the risk of propagating the biases in the generating distribution in the trained models (\citealp{Rosenberg2019SpeechRW}; \citealp{moslem2024leveragingsyntheticaudiodata}).

Gender and age are usually the primary concerns with respect to the representativeness of the dataset. Naturalness in spontaneous speech is often neglected even though it is an equally important factor in creating a dataset that truly represents speech in all its dimensions. Many projects curate data following guidelines that hinder the capture of spontaneous speech, seeking to minimize or prohibit code-switching, background noise, slang, and ungrammatical constructions. Such guidelines reduce the usability of the models for real-world deployment scenarios \cite{diarra2025dealinghardfactslowresource}, as these phenomena reflect the realities of daily interactions and the linguistic evolution of many low resource languages. Day-to-day speech in many African languages feature high rates of inter-sentential and intra-sentential language shifts both between African languages and with high resource colonial languages. Code-switching may enable models to use the high resource language constantly appearing in sentences and conversations to improve accuracy on the LRL.

Among LRLs, predominately-oral languages (POL) constitute a large subset where, speech, to the almost complete exclusion of writing, has been and remains the dominant means of knowledge transmission. Bambara along with most African languages, are POL. For many of those languages, the only readily-available body of natural communication is radio and television broadcasts. This resource includes background and foreground music, various types of noise, and phone calls in which the audience jumps into the conversation, bringing a variety of accents and dialects \cite{doumbouya2021usingradio}. This abundant source of data is rarely exploited due to the many challenges in transcribing such unpredictable conversations and because it implies renouncing control over topics and audio quality. 

In this paper, we introduce \textbf{Kunkado}, a $160$ hour transcribed ASR dataset compiled from radio archives. The title of this paper, \textit{Kunnafonidilaw ka cadeau} can be translated as "Media's Gift". It is also a good example of the everyday code-switching in Bambara, the word \textit{cadeau} being a direct borrowing from French. The entire dataset was automatically transcribed, with 25\% corrected by humans. We report on human evaluation performed on a model trained with the reviewed subset of Kunkado, comparing this result to that of the same model trained with a much larger quantity of curated data. (see section \ref{sec:results}). In the next section, we share insights on handling code-switching and numbers and how trade-offs can be made between consistency and simplicity in transcription.

\section{Characteristics of Kunkado}
\label{sec:dataset}

\begin{table*}[h!]
    \centering
    \begin{tabular}{|c|c|}
        \hline
        \textbf{Tag} & \textbf{Meaning} \\
        \hline \hline
        \textless BRUITS\textgreater & generic noise \\
        \hline
        \textless INCOMPRÉHENSIBLE\textgreater & fully inaudible speech \\
        \hline
        \textless CHEVAUCHEMENT\textgreater & speaker overlap \\
        \hline
        \textless RIRES\textgreater & laughter \\
        \hline
        \textless MUSIQUE\textgreater & music / jingle (no lyrics) \\
        \hline
        \textless TOUX\textgreater & cough \\
        \hline
        \textless INVOCATION\textgreater & prayers, quranic excerpts \\
        \hline
        \textless ECHO\textgreater & echo artifact \\
        \hline
        \textless APPLAUDISSEMENTS\textgreater & applause \\
        \hline
        \textless CRIS\textgreater & screams \\
        \hline
        \textless PLEURES\textgreater & crying \\
        \hline
        \_\_phrase\_\_ & double underscores delimit code-switched words and sentences\\
        \hline
        $\cdots$ & used to mark speech cut-offs and hesitations\\
        \hline
    \end{tabular}
    \caption{Tags/Annotations for acoustic and linguistic events captured in Kunkado}
    \label{tab:tags}
\end{table*}

\paragraph{Audio collection \& segmentation:}We obtained approximately 300 hours of broadcast recordings from 4 Malian radio stations. After segmentation, We only retained segments between 600 milliseconds and 45 seconds of duration. We have released, on HuggingFace\footnote{The full dataset is released under cc-by-sa-4.0 license at \href{https://huggingface.co/datasets/RobotsMali/kunkado}{RobotsMali/kunkado} on HF}, 118,925 segments totaling $161.15$ hours. Approximately 94\% of segments are less than 15 seconds in duration. The mean duration of segments is 4.9 seconds.

For audio segmentation, we employed a simple energy threshold based method implemented via the \texttt{split\_on\_silence} function from the pydub library \cite{robert2018pydub}. This function performs segmentation by analyzing the root mean square (RMS) energy of the audio and splitting the signal where the energy drops below a predefined absolute loudness threshold. Specifically, we used the parameters $min\_silence\_len=600$ milliseconds and $silence\_thresh=-35 \text{ dBFS}$. This configuration ensured that the audio stream was only split when the signal's loudness dropped below $-35 \text{ dB}$ relative to the maximum possible volume for a minimum duration of $600$ milliseconds.

While this kind of silence proxy segmentation is much faster than modern voice-activity-detection based methods and does not discard any part of the original recording---i.e, it simply finds endpoints to split on based on the given parameters, yielding $segments\_duration=original\_duration$, it also results in much rougher segmentation and speech cut-offs. We modelled those cut-offs during transcription (see Table \ref{tab:tags}). 

\begin{figure}[h!]
    \centering
    \includegraphics[width=\linewidth]{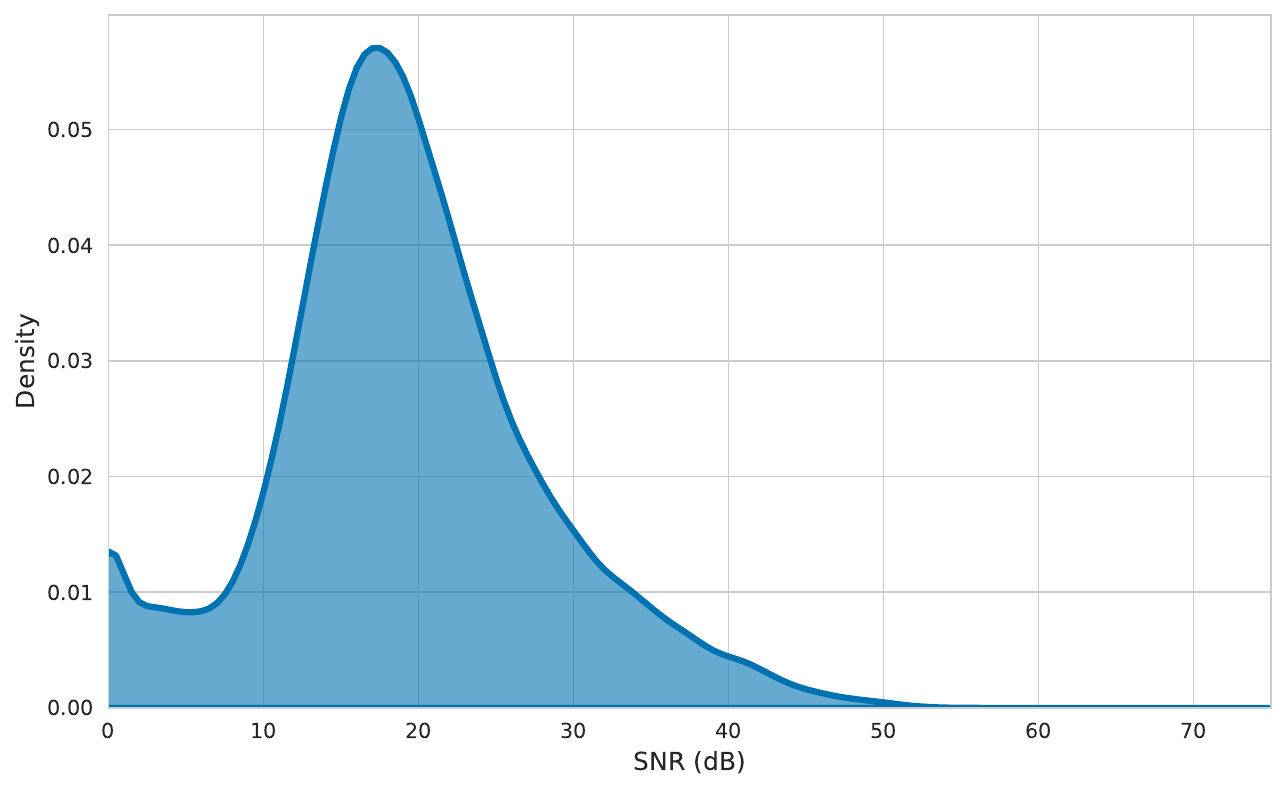}
    \caption{Density Distribution of Signal-to-Noise Ratio values in Kunakdo. This figure includes both subsets}
    \label{fig:snr}
\end{figure}

\paragraph{Noise estimation:}We calculate signal-to-noise ratio (SNR) as an estimate of the level of noise/non-speech signal in the segmented dataset, using the same implementation and classification thresholds defined by \citeauthor{diarra2025dealinghardfactslowresource}. \textbf{69.2}\% of the segments fall above the High SNR category (\textgreater 15 dB). Although they feature a considerable amount of acoustic non-speech information, this relatively good SNR measurement is explained by the fact that the volume of those events is significantly lower than that of speech, as broadcasts are recorded with professional equipment. The measurements demonstrate that radio data is a quality source to create speech datasets \cite{doumbouya2021usingradio}. Figure \ref{fig:snr} shows the density distribution of SNR values in the dataset.

\paragraph{Transcription:} We have re-engineered transcription, aiming at better matching the requirements of the task to the skill level of annotators that were available to us and speeding up the process. We used the process described in  \citeauthor{diarra2025dealinghardfactslowresource}, redefining segment annotation as a first review task where the objective is to correct an automatic transcription generated by \texttt{RobotsMali/soloni-114m-tdt-ctc-v0} Bambara ASR model and to ensure it conforms to our guidelines. 

Our guidelines direct annotators to transcribe numbers with Hindu-Arabic numerals, in contrast to orthodox ASR practice where numbers are written out as words. Writing numerals speeds up transcription, reduces potential ambiguity in parsing the quantity, simplifies downstream processing, and is more easily validated by human reviewers. We also have created data for comparison of number formatting approaches in end-to-end ASR---which remains a challenge even in high resource languages \cite{huber2025handlingnumericexpressionsautomatic} Annotators were instructed to use 13 tags to capture as much acoustic and linguistic information as possible. Code-switching to French was written out using the French orthography, ignoring the existence of Bambara-ized spellings commonly, but inconsistently, used for many French words and expressions. Code-switching to Arabic also followed this principle, though transliterated to the Latin alphabet. A transliteration standard was not enforced due to the specialized knowledge required and the complexity that this would have added to the annotation task. While French is often woven into Bambara in a wide variety of ways, Arabic is generally limited to a small set of formulaic Islamic expressions, though these are used frequently. Table \ref{tab:tags} presents the complete list of tags and their significations. Other rules on proper nouns, acronyms and spelled-out words are similar to those used in the ANV Bambara project \cite{diarra2025dealinghardfactslowresource}. Although annotators were not required to follow a single standard Bambara orthography, they used the Bamadaba dictionary \cite{vydrin:halshs-03909864} as their primary reference. The annatoators used the same data annotation interface described in \cite{diarra2025dealinghardfactslowresource}.

Despite added complexity with respect to tagging and code-switching, our team of seven annotators were able to correct 39.3 hours of segments in roughly 1260 hours of human labor, yielding a $32$x ratio, a bit faster than the 36x reference datapoint reported in the transcription cost analysis study by \cite{diarra2025costanalysishumancorrectedtranscription} which used the same model for generating the automated transcriptions. We speculate that more flexible orthography and the use of numerals rather than spelled-out numbers contributed to this 4-hour difference.

\section{ASR Experiments}
\label{sec:asr}

\begin{table*}[!h]
    \centering
    \begin{tabular}{l@{\hskip 2cm}cccc} 
        \toprule
        \multirow{2}{*}{\textbf{Model}} & \multicolumn{2}{c}{\textbf{WER (\%) $\downarrow$}} & \multicolumn{2}{c}{\textbf{CER (\%) $\downarrow$}} \\
        \cmidrule(lr){2-3} \cmidrule(lr){4-5} 
        
        & \textbf{Kunkado Test} & \textbf{Nyana-Eval} & \textbf{Kunkado Test} & \textbf{{Nyana-Eval}} \\
        \midrule
        
        \multicolumn{5}{l}{\textbf{soloni (jeli-asr)}} \\
        \addlinespace[0.5ex]
        Unfinetuned (v0) & 46.91 & 40.75 & 30.56 & 24.71 \\
        Finetuned (v1) & 39.13 & 39.44 & 20.98 & 20.5 \\
        \addlinespace[1ex]

        \multicolumn{5}{l}{\textbf{soloni (afvoices)}} \\
        \addlinespace[0.5ex]
        Unfinetuned (v2) & 44.47 & 36.07 & 29.61 & 20.24 \\
        Finetuned (v3) & 37.12 & \textbf{32.33} & 21.17 & \textbf{16.72} \\
        \addlinespace[1ex]
        
        \bottomrule
    \end{tabular}
    \caption{ASR Evaluation results: We apply the same normalization steps as explained in section \ref{sec:asr} and remove the tags from both the reference and the prediction before calculating the WER and CER. The values in bold highlight the best performances per metric across both benchmarks}
    \label{tab:metrics}
\end{table*}

We finetuned multiple Bambara ASR models, previously trained on Jeli-ASR \cite{Diarra2022Griots}, from RobotsMali's baseline ASR experiments. These models are based on NVIDIA's Parakeet family of monolingual English ASR models \cite{rekesh2023fastconformerlinearlyscalable}. We evaluated all the models on a 5 hour test set taken from the Kunkado data, and Nyana-Eval, a small, stratified human evaluation dataset with only 45 entries of 3 minutes total duration \cite{diarra2025dealinghardfactslowresource}. We report in section \ref{sec:results} the automatic and human evaluation results for all the RobotsMali soloni models, except the v3 version for which we only report the WER gains since it was not part of the human evaluation\footnote{We still release all the other models with the corresponding WER scores. \href{https://hf.co/RobotsMali/models}{hf.co/RobotsMali/models}}.

\paragraph{Experimental setup:} We used 4 NVIDIA A100 GPUs with 80GB VRAM each for these experiments. We finetuned \texttt{soloni-114m-tdt-ctc-v0} and  \texttt{soloni-114m-tdt-ctc-v2} in this study. The two models have an hybrid architecture with a Fast-Conformer encoder \cite{rekesh2023fastconformerlinearlyscalable} and two jointly trained decoders: an autoregressive TDT decoder, Token-and-Duration Transducer (\citealp{xu2023efficientsequencetransductionjointly}; \citealp{graves2012sequencetransductionrecurrentneural}) and a convolutional decoder trained with a Connectionist Temporal Classification (CTC) loss function \cite{gravesctc2006}. We will refer to the models as soloni-v0 and soloni-v2. The finetuned versions of these models are identified on HuggingFace as v1 and v3, respectively.

\paragraph{Training Data:} We finetuned the two models on the training set of the human-reviewed subset of Kunkado (33.47 hours). We simplified training for the ASR model by removing symbols and diacritics, and making number and Bambara-specific normalizations using our \href{https://pypi.org/project/bambara-normalizer/}{bambara-normalizer} Python package. We then removed code-switching markers (the double underscores) and all punctuation. We kept only a reduced set of acoustic event tags (e.g., overlaps, paralinguistic vocalizations such as laughter, and music), which are modeled during training but ignored during evaluation. We applied these normalization steps after noticing that, with only just over 30 hours of training data, the models were struggling with the human-annotation variability present in the reference transcripts. These models rarely improved beyond 70\% WER due to inconsistencies in numbers, tags and code-switching. We concluded that it would require much more data for an end-to-end ASR model to learn the original task. Table \ref{tab:normalize} contains 3 examples of how this normalization simplifies the task.

\begin{table*}
    \centering
    \begin{tabularx}{\textwidth}{p{0.45\textwidth}|X}
        Original & Normalized \\
        \toprule
        Bamananw ko ten ko maa jugu t'i ba sinamuso ye, nka n'a ni ba b\textipa{E} k\textipa{E}l\textipa{E} la, \textless CHEVAUCHEMENT\textgreater \_\_ voilà\_\_ o k\textipa{O}ni ka di ye \_\_ donc\_\_ jamana... & \textcolor{green}{b}amananw ko ten ko maa jugu t'i ba sinamuso ye nka n'a ni ba b\textipa{E} k\textipa{E}l\textipa{E} la \textcolor{green}{\textless tag\textgreater} voilà o k\textipa{O}ni ka di ye donc jamana\\
        \hline
        \textless MUSIQUE\textgreater an b'an sinsin ni Alahutala t\textipa{O}g\textipa{O} barikama ye. & \textcolor{green}{\textless MUSIQUE\textgreater} an b'an sinsin ni Alahutala t\textipa{O}g\textipa{O} barikama ye\\
        \hline
        \textipa{EE}... n\textipa{E}g\textipa{E}juru sira 76 64 10 10... \_\_76 64 10 10\_\_ & \textipa{EE} n\textipa{E}g\textipa{E}juru sira \textcolor{green}{bi wolonwula ni w\textipa{OO}r\textipa{O} bi w\textipa{OO}r\textipa{O} ni naani tan tan soixante-seize soixante-quatre dix dix}\\
        \bottomrule
    \end{tabularx}
    \caption{Sample Kunkado transcripts pre and post normalization}
    \label{tab:normalize}
\end{table*}

\paragraph{Training configurations:} We trained soloni-v0 for 100k steps on 2 GPUs, with batch size 40, using the AdamW optimizer and Noam scheduler with a learning rate scaling factor of $0.03$ and a 10\% warmup ratio \cite{vaswani2023attentionneed}. We froze the encoder for the last 7,000 steps, training only the 5M combined parameters of the two decoders. 

For soloni-v2, trained on much more data with the ANV dataset, we simply trained all the 114M parameters for 13k steps on 4 GPUs, with batch size of 64 and LR scaling factor of 1 and 3,000 warmup steps. Both models were trained with \texttt{bf16} float precision.

\section{Evaluation \& Results}
\label{sec:results}

We evaluated the Word Error Rate (WER) and Character Error Rate (CER) of the resulting models, and we also report the findings of the human evaluation conducted during the Bambara ANV project (\citealp{diarra2025dealinghardfactslowresource}, \citealp{tall_2025_17672774}). Table \ref{tab:metrics} presents the WERs and CERs of the two models before and after our finetuning experiments; the terms in parentheses represent the dataset on which the unfinetuned versions were trained and their version IDs on HuggingFace. Since these models have two decoders, we only report the best scores; detailed per-decoder metrics can be found in their respective model cards. Both finetuned versions reduce WER on the Kunkado test set. Soloni-v3, our latest finetuned model, achieves the best results on both benchmarks\footnote{Note that, as one third of Nyana-Eval (15 examples) comes from the Kunkado test set, there is a small intersection between the two test sets}.

\begin{figure}[h!]
    \centering
    \includegraphics[width=\linewidth]{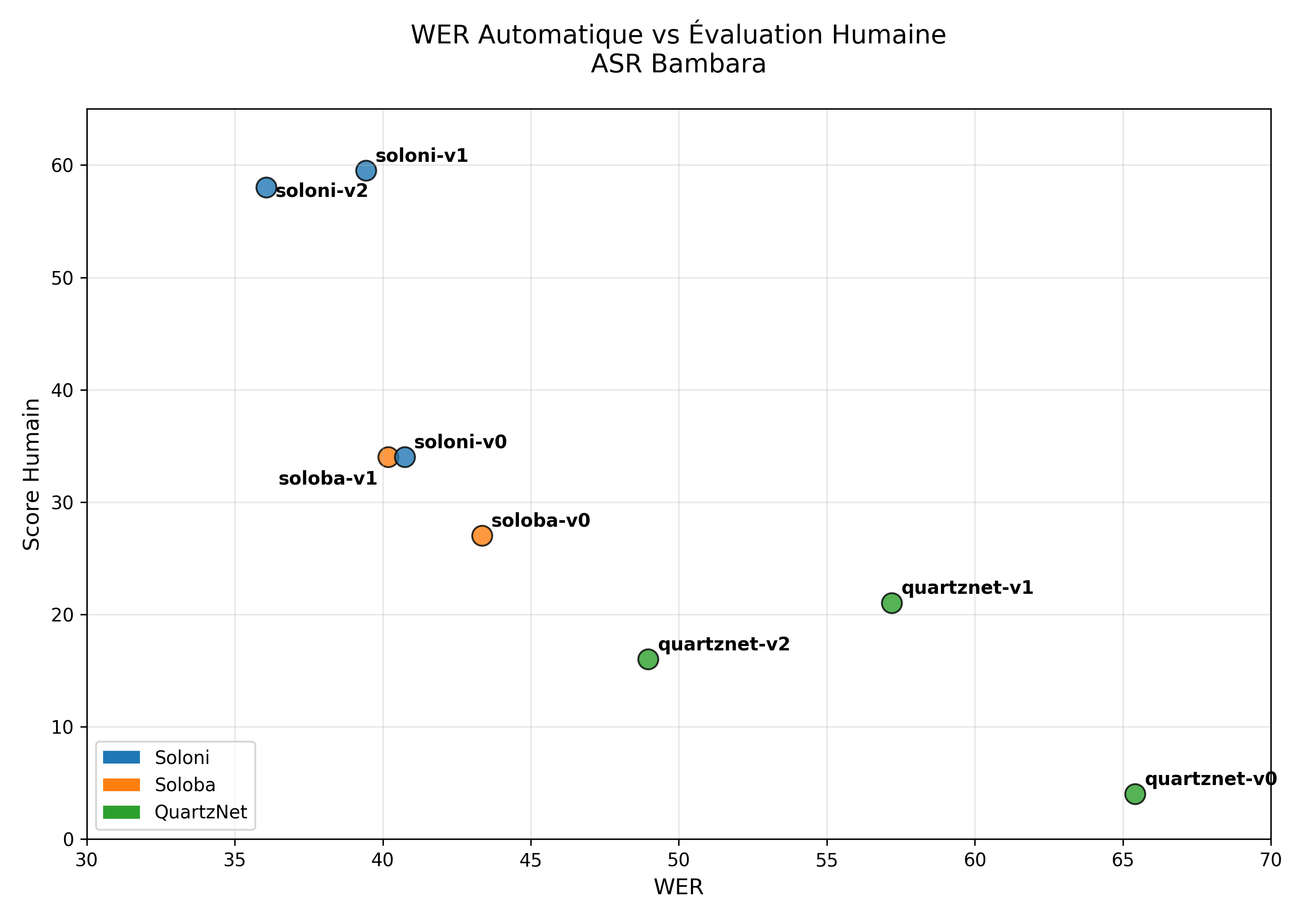}
    \caption{WER vs human evaluation. Figure from \cite{tall_2025_17672774}}
    \label{fig:human-eval}
\end{figure}

In the detailed comparative analysis report by \citeauthor{tall_2025_17672774} on several RobotsMali ASR models, we found that soloni-v1 achieved the highest human evaluation score on Nyana-Eval (59.5 out of 135), outperforming soloni-v2, which was trained with an additional 98 hours from the African Next Voices Bambara dataset. According to \citeauthor{tall_2025_17672774}, soloni-v1 shows clear gains in handling speech disfluencies, proper names, code-switching, and noisy or overlapping speech—phenomena that are common in natural Bambara conversations \cite{tall_2025_17672774}. The v3 models are not included in that report because they were fine-tuned from v2 after the report was published; however, the substantial WER improvement in soloni-v3 (32.33\% vs. 36.07\%) suggests that a proportional improvement in human scores is plausible. Figure \ref{fig:human-eval} (from the comparative analysis report) ranks the models by both human scores and WER. All of these models, along with their v3 variants, are available on RobotsMali’s HuggingFace profile.

\section{Conclusion}
\label{sec:conclusion}
In this work, we introduce Kunkado, a 160-hour Bambara ASR dataset compiled from Malian radio archives, "Media's Gift" to LRL Bambara NLP, and designed to better reflect present-day, naturally occurring speech. By shifting from curated source materials and controlled recording conditions toward broadcast content containing code-switching, disfluencies, background noise, and overlapping speech, we address a major source of domain mismatch that often yields models that perform well on curated data using standard metrics but poorly on real-world data and applications. Our experiments show that finetuning on 33.47 hours of human-reviewed Kunkado data yields substantial gains, and the best-performing configuration (soloni-v3) improves WER on both Kunkado-sourced test data and Nyana-Eval relative to its unfinetuned counterpart. These results support the broader conclusion that, for Bambara, representativeness and linguistic realism can matter as much as (or more than) raw hours when the goal is real-world usability. At the same time, the linguistic richness of spontaneous speech increases annotation and modeling difficulty, motivating pragmatic transcription guidelines and normalization choices, as well as continued investment in human review. We release Kunkado and the associated models to encourage research, community-driven quality standards led by native speakers, and future work on code-switching-aware evaluation and data collection and training from large-scale resources such as radio broadcasts for low-resource, predominantly oral languages.

\section{Limitations}
\label{sec:limitations}
The design of the Kunkado corpus intentionally prioritized the linguistic authenticity of the Bambara language. Specifically, the data reflects the most common and contemporary register of spoken Bambara, characterized by features inherent to natural conversational settings (e.g., podcasts, TV shows, and debates). These features include extensive code-switching, prevalence of slang, and frequent spontaneous speech disfluencies.

Our ASR experiments demonstrated that these added linguistic complexities necessitate significantly greater volumes of data and advanced model engineering to achieve robust performance. Although radio broadcasts, in the West African region, represent a virtually inexhaustible, daily-generated source of data, resource constraints limited our labeling efforts to only $\approx 40$ hours within the scope of this project, relying solely on limited internal funding.

The compelling results from the human evaluation indicate that continued annotation and development on more authentic data could substantially accelerate the deployment of high-fidelity, real-world speech applications for Bambara speakers. Moving forward, we advocate for a community-driven dataset design strategy where quality standards are organically defined with robuts participation by native-speakers.

\section{Acknowledgments}
We would like to extend our gratitude to Radio Benkouma, Mousso TV, ORTM and Radio Sahel FM for graciously sharing their archives with us and allowing us to release the data with an open source license.

\bibliography{custom}

@INPROCEEDINGS{cmuwilderness_black,
  author={Black, Alan W},
  booktitle={ICASSP 2019 - 2019 IEEE International Conference on Acoustics, Speech and Signal Processing (ICASSP)}, 
  title={CMU Wilderness Multilingual Speech Dataset}, 
  year={2019},
  volume={},
  number={},
  pages={5971-5975},
  doi={10.1109/ICASSP.2019.8683536}
}

@misc{Diarra2022Griots,
  author                = {Sebastien Diarra and Michael Leventhal and Allahsera Auguste Tapo},
  title                 = {RobotsMali Griots Speech Dataset, and ASR},
  howpublished          = {\url{https://github.com/robotsmali-ai/jeli-asr/}},
  year                  = {2022},
}

@misc{pratap2023scalingspeechtechnology1000,
      title={Scaling Speech Technology to 1,000+ Languages}, 
      author={Vineel Pratap and Andros Tjandra and Bowen Shi and Paden Tomasello and Arun Babu and Sayani Kundu and Ali Elkahky and Zhaoheng Ni and Apoorv Vyas and Maryam Fazel-Zarandi and Alexei Baevski and Yossi Adi and Xiaohui Zhang and Wei-Ning Hsu and Alexis Conneau and Michael Auli},
      year={2023},
      eprint={2305.13516},
      archivePrefix={arXiv},
      primaryClass={cs.CL},
      url={https://arxiv.org/abs/2305.13516}, 
}

@misc{xu2023efficientsequencetransductionjointly,
      title={Efficient Sequence Transduction by Jointly Predicting Tokens and Durations}, 
      author={Hainan Xu and Fei Jia and Somshubra Majumdar and He Huang and Shinji Watanabe and Boris Ginsburg},
      year={2023},
      eprint={2304.06795},
      archivePrefix={arXiv},
      primaryClass={eess.AS},
      url={https://arxiv.org/abs/2304.06795}, 
}

@misc{rekesh2023fastconformerlinearlyscalable,
      title={Fast Conformer with Linearly Scalable Attention for Efficient Speech Recognition}, 
      author={Dima Rekesh and Nithin Rao Koluguri and Samuel Kriman and Somshubra Majumdar and Vahid Noroozi and He Huang and Oleksii Hrinchuk and Krishna Puvvada and Ankur Kumar and Jagadeesh Balam and Boris Ginsburg},
      year={2023},
      eprint={2305.05084},
      archivePrefix={arXiv},
      primaryClass={eess.AS},
      url={https://arxiv.org/abs/2305.05084}, 
}

@misc{vaswani2023attentionneed,
      title={Attention Is All You Need}, 
      author={Ashish Vaswani and Noam Shazeer and Niki Parmar and Jakob Uszkoreit and Llion Jones and Aidan N. Gomez and Lukasz Kaiser and Illia Polosukhin},
      year={2017},
      eprint={1706.03762},
      archivePrefix={arXiv},
      primaryClass={cs.CL},
      url={https://arxiv.org/abs/1706.03762}, 
}

@misc{graves2012sequencetransductionrecurrentneural,
      title={Sequence Transduction with Recurrent Neural Networks}, 
      author={Alex Graves},
      year={2012},
      eprint={1211.3711},
      archivePrefix={arXiv},
      primaryClass={cs.NE},
      url={https://arxiv.org/abs/1211.3711}, 
}

@inproceedings{gravesctc2006,
author = {Graves, Alex and Fernández, Santiago and Gomez, Faustino and Schmidhuber, Jürgen},
year = {2006},
month = {01},
pages = {369-376},
title = {Connectionist temporal classification: Labelling unsegmented sequence data with recurrent neural 'networks},
volume = {2006},
booktitle = {ICML 2006},
journal = {ICML 2006 - Proceedings of the 23rd International Conference on Machine Learning},
doi = {10.1145/1143844.1143891}
}

@article{vydrin:halshs-03909864,
  TITLE = {{Vers un dictionnaire orthographique bambara}},
  AUTHOR = {Vydrin, Valentin Feodosievich},
  URL = {https://shs.hal.science/halshs-03909864},
  JOURNAL = {{Mandenkan : Bulletin Semestriel d'{\'E}tudes Linguistiques Mand{\'e}}},
  PUBLISHER = {{Presses de l'Inalco}},
  NUMBER = {68},
  PAGES = {59-82},
  YEAR = {2022},
  MONTH = Dec,
  DOI = {10.4000/mandenkan.2905},
  HAL_ID = {halshs-03909864},
  HAL_VERSION = {v1},
}

@misc{diarra2025costanalysishumancorrectedtranscription,
      title={Cost Analysis of Human-corrected Transcription for Predominately Oral Languages}, 
      author={Yacouba Diarra and Nouhoum Souleymane Coulibaly and Michael Leventhal},
      year={2025},
      eprint={2510.12781},
      archivePrefix={arXiv},
      primaryClass={cs.CL},
      url={https://arxiv.org/abs/2510.12781}, 
}

@misc{tall_2025_17672774,
author = {Tall, Madani Amadou},
title = {Analyse comparative humaine des modèles ASR Bambara de RobotsMali},
month = nov,
year = 2025,
publisher = {Zenodo},
doi = {10.5281/zenodo.17672774},
url = {https://doi.org/10.5281/zenodo.17672774},
}

@misc{diarra2025dealinghardfactslowresource,
      title={Dealing with the Hard Facts of Low-Resource African NLP}, 
      author={Yacouba Diarra and Nouhoum Souleymane Coulibaly and Panga Azazia Kamaté and Madani Amadou Tall and Emmanuel Élisé Koné and Aymane Dembélé and Michael Leventhal},
      year={2025},
      eprint={2511.18557},
      archivePrefix={arXiv},
      primaryClass={cs.CL},
      url={https://arxiv.org/abs/2511.18557}, 
}

@misc{moslem2024leveragingsyntheticaudiodata,
      title={Leveraging Synthetic Audio Data for End-to-End Low-Resource Speech Translation}, 
      author={Yasmin Moslem},
      year={2024},
      eprint={2406.17363},
      archivePrefix={arXiv},
      primaryClass={cs.CL},
      url={https://arxiv.org/abs/2406.17363}, 
}

@article{Rosenberg2019SpeechRW,
  title={Speech Recognition with Augmented Synthesized Speech},
  author={Andrew Rosenberg and Yu Zhang and Bhuvana Ramabhadran and Ye Jia and Pedro J. Moreno and Yonghui Wu and Zelin Wu},
  journal={2019 IEEE Automatic Speech Recognition and Understanding Workshop (ASRU)},
  year={2019},
  pages={996-1002},
  url={https://api.semanticscholar.org/CorpusID:202889273}
}

@misc{derenzi2025syntheticvoicedataautomatic,
      title={Synthetic Voice Data for Automatic Speech Recognition in African Languages}, 
      author={Brian DeRenzi and Anna Dixon and Mohamed Aymane Farhi and Christian Resch},
      year={2025},
      eprint={2507.17578},
      archivePrefix={arXiv},
      primaryClass={cs.CL},
      url={https://arxiv.org/abs/2507.17578}, 
}

@inproceedings{asr-pipeline-for-lrl-pomak,
author = {Tsoukala, Chara and Kritsis, Kosmas and Douros, Ioannis and Kokkas, Nikolaos and Arampatzakis, Vasileios and Sevetlidis, Vasileios and Markantonatou, Stella and Pavlidis, George},
year = {2023},
month = {01},
pages = {40-45},
title = {ASR pipeline for low-resourced languages: A case study on Pomak},
doi = {10.18653/v1/2023.fieldmatters-1.5}
}

@inproceedings{doumbouya2021usingradio,
    title={Using Radio Archives for Low-Resource Speech Recognition: Towards an Intelligent Virtual Assistant for Illiterate Users},
    author={Doumbouya, Moussa and Einstein, Lisa and Piech, Chris},
    booktitle={Proceedings of the AAAI Conference on Artificial Intelligence},
    volume={35},
    year={2021}
  }

@misc{robert2018pydub,
  title={Pydub: Manipulate audio with a simple and easy high level interface},
  author={Robert, James},
  year={2018},
  publisher={GitHub},
  url={http://pydub.com/}
}

@misc{huber2025handlingnumericexpressionsautomatic,
      title={Handling Numeric Expressions in Automatic Speech Recognition}, 
      author={Christian Huber and Alexander Waibel},
      year={2025},
      eprint={2408.00004},
      archivePrefix={arXiv},
      primaryClass={eess.AS},
      url={https://arxiv.org/abs/2408.00004}, 
}

\end{document}